\documentclass[letterpaper, 10 pt, conference]{ieeeconf}  

\IEEEoverridecommandlockouts                              

\usepackage{graphics} 
\usepackage{epsfig} 
\usepackage{mathptmx} 
\usepackage{times} 
\usepackage{amsmath} 
\usepackage{amssymb}  

\usepackage{tabularx}
\usepackage{comment}
\usepackage{pgfplots}
\usepackage{booktabs}
\usepackage{soul} 
\usepackage{url}

\title{\LARGE \bf
A Review of Personalisation in Human-Robot Collaboration and Future Perspectives Towards Industry 5.0
}

\author{James Fant-Male$^{1}$ and Roel Pieters$^{1}$
\thanks{$^{1}$Cognitive Robotics group, Unit of Automation Technology and Mechanical Engineering, Tampere University, 33720, Tampere, Finland;
        {\tt\small firstname.surname@tuni.fi}}%
}

\begin{document}

\maketitle
\thispagestyle{empty}
\pagestyle{empty}

\begin{abstract}
The shift in research focus from Industry 4.0 to Industry 5.0 (I5.0) promises a human-centric workplace, with social and well-being values at the centre of technological implementation. Human-Robot Collaboration (HRC) is a core aspect of I5.0 development, with an increase in adaptive and personalised interactions and behaviours.
This review investigates recent advancements towards personalised HRC, where user-centric adaption is key. There is a growing trend for adaptable HRC research, however there lacks a consistent and unified approach. The review highlights key research trends on which personal factors are considered, workcell and interaction design, and adaptive task completion. This raises various key considerations for future developments, particularly around the ethical and regulatory development of personalised systems, which are discussed in detail.

\end{abstract}

\section{Introduction}
Significant research has been done into Human-Robot Collaboration (HRC) for manufacturing. This has been fuelled by the pursuit of Industry 4.0 (I4.0), incorporating advanced technologies such as robotics, cloud computing, and big data~\cite{xu_industry_2021}. The proposal of Industry 5.0 (I5.0) by the European Commission in 2021 presents a future vision for industry in Europe~\cite{directorate-general_EC_industry50_2021}. The key shift is from a technology-driven to a value-driven focus~\cite{xu_industry_2021}, with three core values of human-centricity, sustainability and resilience~\cite{xu_industry_2021, directorate-general_EC_industry50_2021}.

The I5.0 paradigm focuses on how technology can be developed and implemented in a way that promotes social values, well-being and human-centricity~\cite{xu_industry_2021, dhanda_reviewing_2025}. It is therefore important to review current advances in HRC through the lens of the human-centred I5.0 paradigm. Given the key focus of human-centricity for I5.0, there has been a lack of broad discussion around the personalisation and adaption of HRC to create user-centric interactions. This is key given human-centred research has been identified as essential to the continued sustainability of HRC~\cite{simoes_designing_2022}.

Various research investigates \textit{human}-centric HRC, where the design and implementation of HRC focuses on the requirements of generic humans. However, there is a greater need to focus on \textit{user}-centered HRC, where interactions are adjusted to specific users, not generic people. The difference between human-aware and user-aware systems, and its importance given the highly variable nature of the human workforce has been discussed previously~\cite{umbrico_design_2022}, and is highly linked to the ``one-of-a-kind operator" concept~\cite{directorate-general_EC_industry50_2021}.

Increasing consideration for human factors in HRC can improve the physical and mental well-being of workers, while maintaining or improving productivity\,~\cite{calzavara_multi-objective_2023, faccio_human_2023}. Along with this, the development of HRC in factories presents a key opportunity to include a wider variety of people within the workplace\,~\cite{stohr_adaptive_2018, mandischer_toward_2023}. This includes the elderly, people with disabilities, and those who may not be disabled but have injury or impairments inhibiting work. Including a wider variety of people in the workforce has various benefits for individuals as well as businesses~\cite{rojas_lack_2024}.

This review surveys the current state of the art related to personalisation of HRC in manufacturing. Various key avenues of research have been pursued in isolation, however there lacks an overarching consensus and discussion around what personalised HRC in manufacturing should entail. It is hoped that this work will bring to light this key aspect of I5.0 and lead to greater discussion and consolidation of efforts to create a unified approach. Various recent surveys have investigated aspects of HRC, however none with the same focus on personalisation considering the I5.0 paradigm. Key related reviews are highlighted in Table\,\ref{tab:surveys}.

\begin{table}[b]
    \centering
    \caption{Related review papers on HRC.}
    \vspace{-0.2cm}
    \begin{tabularx}{\linewidth}{ccX}
        \toprule
        \textbf{Ref.} & \textbf{Year} & \textbf{Review Focus} \\
        \midrule
        \cite{dhanda_reviewing_2025} & 2025 & HRC in I5.0 and future prospects. \\
        \cite{simoes_designing_2022} & 2022 & Review of HRC workspace design. \\
        \cite{faccio_human_2023} & 2023 & Trends of human factors and cobots in manufacturing. \\
        \cite{rojas_lack_2024} & 2024 & Disabled worker assistive technology lack of validation. \\
        \cite{vermeulen_safety_2024} & 2024 & Human factor analysis across HRC. \\
        \cite{li_trustworthy_2025} & 2025 & Trustworthy AI for human-centric smart manufacturing. \\ 
        \cite{gasteiger_factors_2023} & 2023 & Personalisation factors for HRI review. \\
        \cite{gaffinet_human_2025} & 2025 & Human Digital Twins in I5.0. \\ 
        \cite{ramos_collaborative_2024} & 2024 & Collaborative intelligence for safety in industry. \\
        \bottomrule
    \end{tabularx}
    \label{tab:surveys}
\end{table}

The remainder of this review is structured as follows.
Section\,\ref{sec:methods} presents the search methodology and overall research trends. Section\,\ref{sec:variability} gives an overview of human factors used for personalised HRC and Human Digital Twins (HDT). Section\,\ref{sec:workcell_interface} discusses the design of HRC workcells and interfaces. Section\,\ref{sec:task_completion} investigates user-robot dynamics in collaborative task completion. Section\,\ref{sec:ethics} presents regulatory and ethical considerations. Section\,\ref{sec:future_work} gives a discussion on future developments and Section\,\ref{sec:conclusion} presents final conclusions.

\section{Methodology and Overall Trends}\label{sec:methods}
The review is conducted following a search of key literature databases, namely Web of Science, Scopus and IEEE eXplore. The search term is designed to encompass relevant works on HRC in industry with a personalised or adaptive nature. Only peer-reviewed conference and journal papers are considered, which must be available in English and published since 2018. The search is performed on 17\textsuperscript{th} January 2025 across titles, abstracts and keywords using the search term:

\textit{(human-robot collaboration OR HRC) AND (manufacturing OR industry OR workplace) AND (personal* OR custom* OR adapt* OR user-centric OR human-centric)}

\begin{figure}[t]
    \centering
    \includegraphics[width=\linewidth]{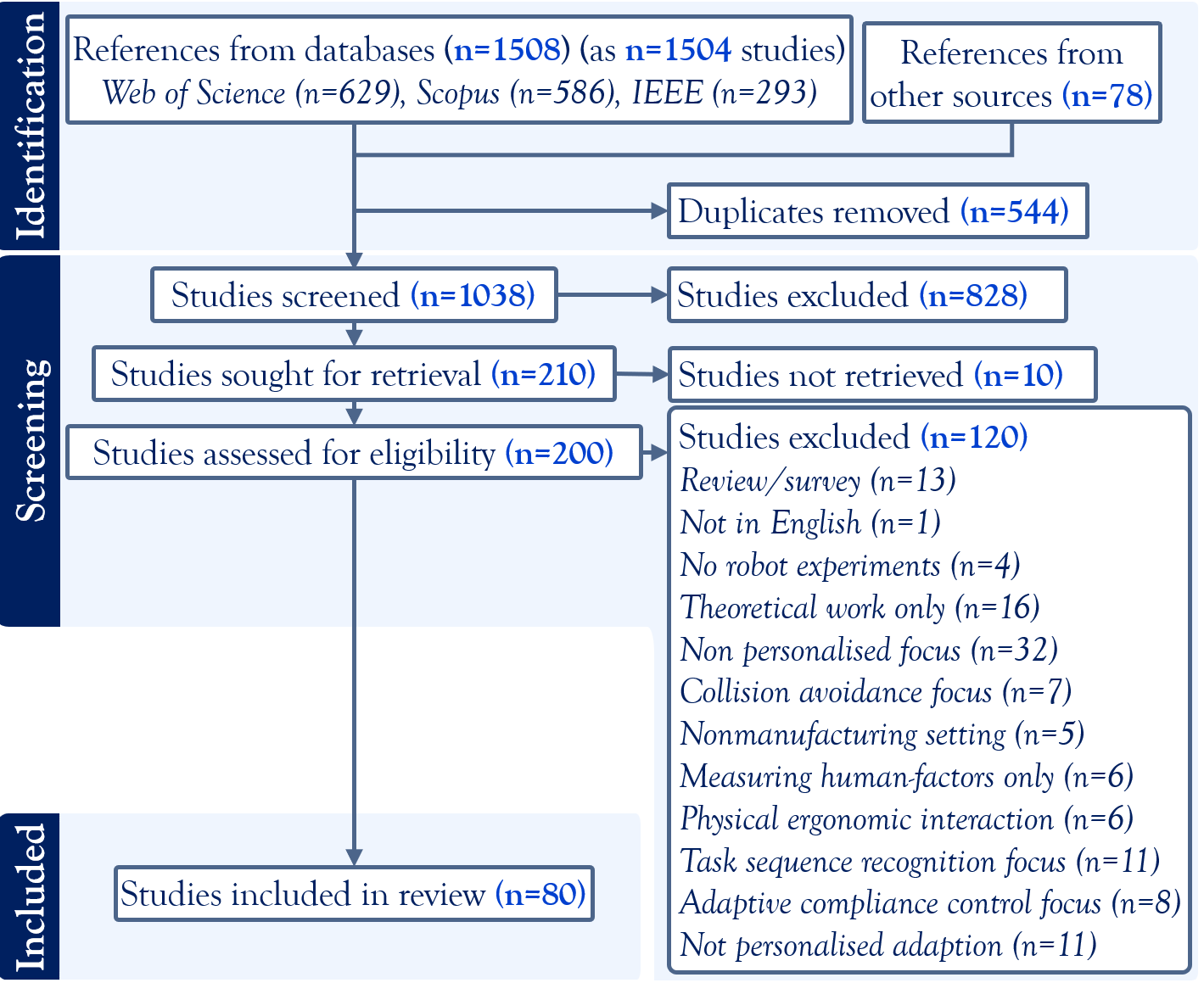}
    \vspace{-0.6cm}
    \caption{PRISMA Flow Diagram of Review Process.}
    \vspace{-0.5cm}
    \label{fig:prisma}
\end{figure}

The PRISMA analysis framework is followed using Covidence software, illustrated in Fig.\,\ref{fig:prisma}. The initial search yielded 1508 results, with an additional 78 included through snowballing and additional searching. 544 duplicates are removed, 4 merged as the same study and 828 removed as irrelevant during title and abstract screening. 200 are accessible and go through full-text review of which 120 are excluded. The primary reasons for exclusion are: not having a personalised focus/adapting to generic humans; review papers; purely theoretical focus. As this review is interested in high-level behaviour adaption, works focussing on low-level control aspects such as adaptive compliance control, adaptive ergonomic handover positioning or collision avoidance are excluded. After all exclusions 80 works\footnote{Available at: \url{https://doi.org/10.5281/zenodo.14930099}} are considered in the subsequent analysis.

\begin{figure}[tb]
    \centering
    \includegraphics[width=1\linewidth]{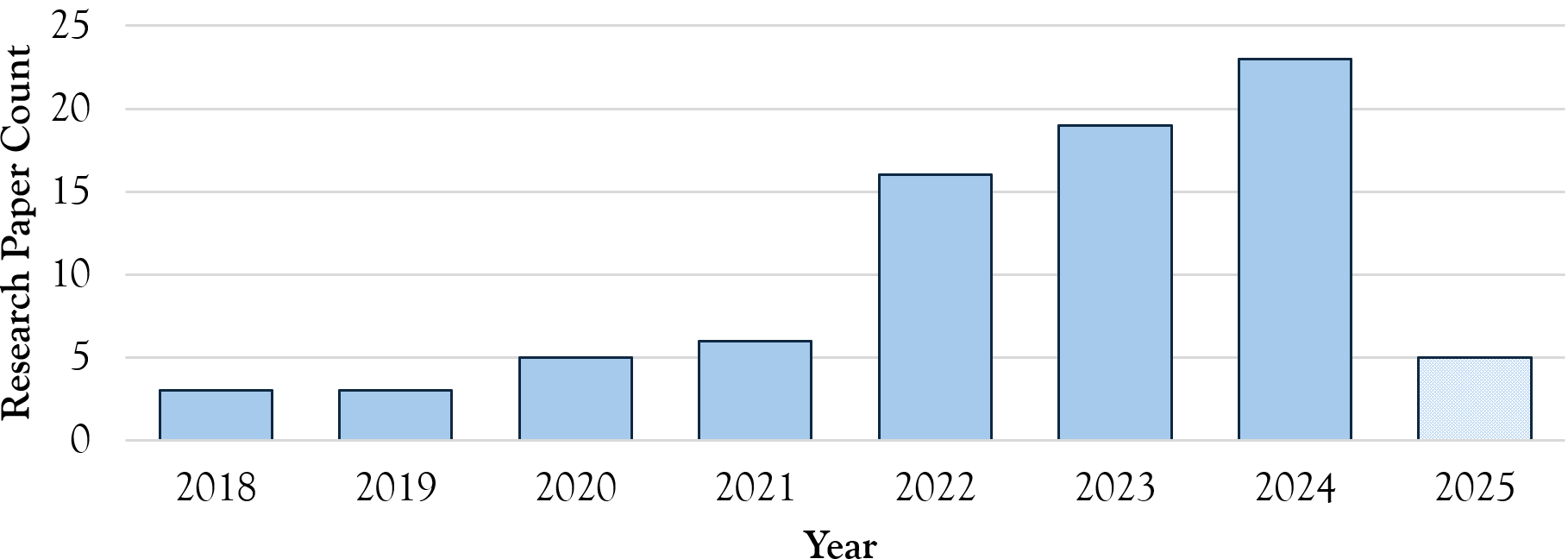}
    \vspace{-0.6cm}
    \caption{Analysis of publication count by year.}
    \label{fig:years}
\end{figure}

Analysis of the research output by year is shown in Fig.\,\ref{fig:years}. The increase year-on-year is evident, with a particular jump since 2021. This may be attributed to the EU's shift in focus towards user-centric workplaces in line with the 2021 I5.0 proposals~\cite{directorate-general_EC_industry50_2021}, and verifies the need for a review of this growing research area. The locations of contributing institutions illustrated in Table\,\ref{tab:country_counts} highlights significant research outputs in Europe, particularly Italy, the UK and Germany. 

\begin{table}[tb]
\centering
\caption{Publication count by country.}
\vspace{-0.2cm}
\newcommand\x{ }
\begin{tabular}{@{\x}c@{\x}c@{\x}@{\x}c@{\x}c@{\x}@{\x}c@{\x}c@{\x}}
\toprule
Country & Count & Country & Count & Country & Count \\
\midrule
Italy & 27 & India & 2 & South Korea & 1 \\
Germany & 16 & Spain & 2 & Belgium & 1 \\
UK & 13 & Sweden & 2 & Columbia & 1 \\
US & 12 & Portugal & 2 & Mexico & 1 \\
China & 5 & Finland & 2 & Switzerland & 1 \\
Denmark & 4 & France & 2 & Cyprus & 1 \\
Netherlands & 4 & Slovenia & 2 & Israel & 1 \\
Greece & 3 & South Africa & 2 & Luxembourg & 1 \\
New Zealand & 3 & Austria & 1 & Singapore & 1 \\
\bottomrule
\end{tabular}
\vspace{-0.4cm}
\label{tab:country_counts}
\end{table}

\begin{figure}[tb]
    \centering
    \vspace{-0.2cm}
    \includegraphics[width=1\linewidth]{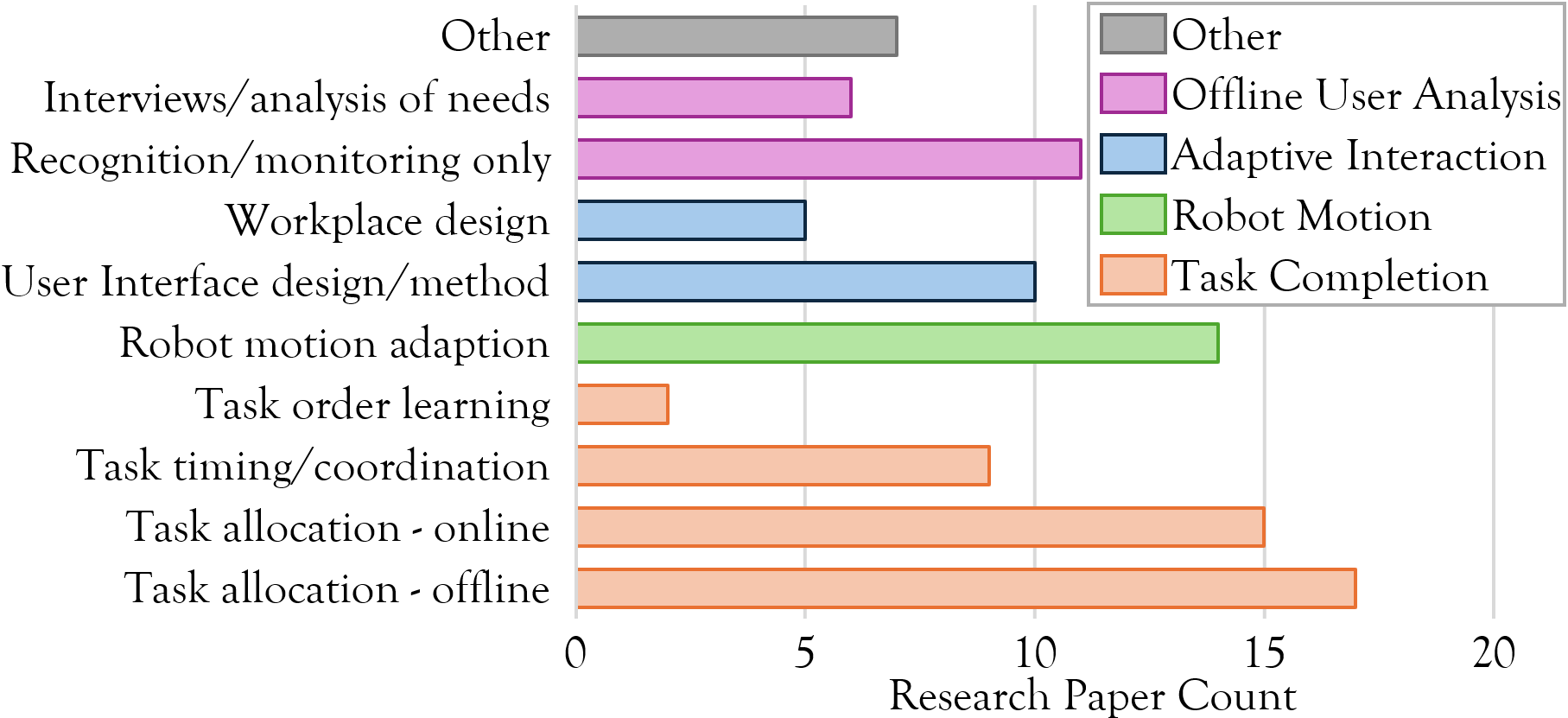}
    \vspace{-0.6cm}
    \caption{Article research focus areas \textit{(Articles may fit in multiple categories)}.}
    \label{fig:focus_area}
\end{figure}

Articles are categorised according to manually identified focus areas on how HRC is personalised. Representative works from each category are highlighted in Table\,\ref{tab:rep_papers}. Subsequent grouping identified four overarching themes, shown in Fig.\,\ref{fig:focus_area}.  These themes have been used to structure this review, with \textit{Adaptive Interaction} and \textit{Task Completion} discussed in Sections~\ref{sec:workcell_interface} and~\ref{sec:task_completion} respectively. Studies in the remaining themes are best compared in relation to the range of human factors used for personalisation, discussed in Section~\ref{sec:variability}.

\begin{table*}[tb]
\centering
\caption{Representative papers of the research focus areas identified.}
\vspace{-0.2cm}
\newcommand\x{ }
\begin{tabularx}{\linewidth}{clX}
\toprule
Ref. & Focus Area & Summary \\
\midrule
\cite{fletcher_adaptive_2020} & Interviews/analysis of needs & Mixed-methods analysis of requirements for adaptive automation assembly. \\
\cite{lagomarsino_pick_2023} & Recognition/monitoring only & Method for online monitoring of cognitive load under different robot collaboration scenarios. \\
\cite{lottermoser_method_2024} & Workplace design & Method for selection of HRC interface technology based on needs of physically challenged user. \\
\cite{andronas_multi-modal_2021} & User Interface design/method & User profile enables custom information display and features on tablet, AR and watch interfaces. \\
\cite{campagna_promoting_2024} & Robot motion adaption & Adjust velocity, separation distance and proximity to user based on user's preference to promote trust. \\
\cite{obidat_development_2023} & Task order learning & Robot dynamically learns human preference for task order and adapts action order accordingly. \\
\cite{lee_bayesian_2024} & Task timing/coordination & Adapt robot task timing based on user proficiency to help balance assembly line. \\
\cite{calzavara_achieving_2024} & Task allocation - online & Dynamic task allocation between user and robot considering energy expenditure of user and makespan. \\
\cite{calzavara_multi-objective_2023} & Task allocation - offline & Task allocation with optimisation for makespan, operator’s energy expenditure and average mental workload. \\
\bottomrule
\end{tabularx}
\vspace{-0.4cm}
\label{tab:rep_papers}
\end{table*}

\section{Variability in the Workforce}\label{sec:variability}
At the core of reframing ``human-centric" design as ``user-centric" design is the awareness of a diverse workforce. Different users must interact with robotic systems, and they will have different physical and mental attributes. The desire to create work environments sympathetic to user variability and inclusion is well documented within the I5.0 paradigm~\cite{xu_industry_2021, directorate-general_EC_industry50_2021}. Understanding the different human factors influencing interactions between user and robot is necessary to properly implement personalised HRC experiences.

\begin{figure}[tb]
    \centering
    \includegraphics[width=1\linewidth]{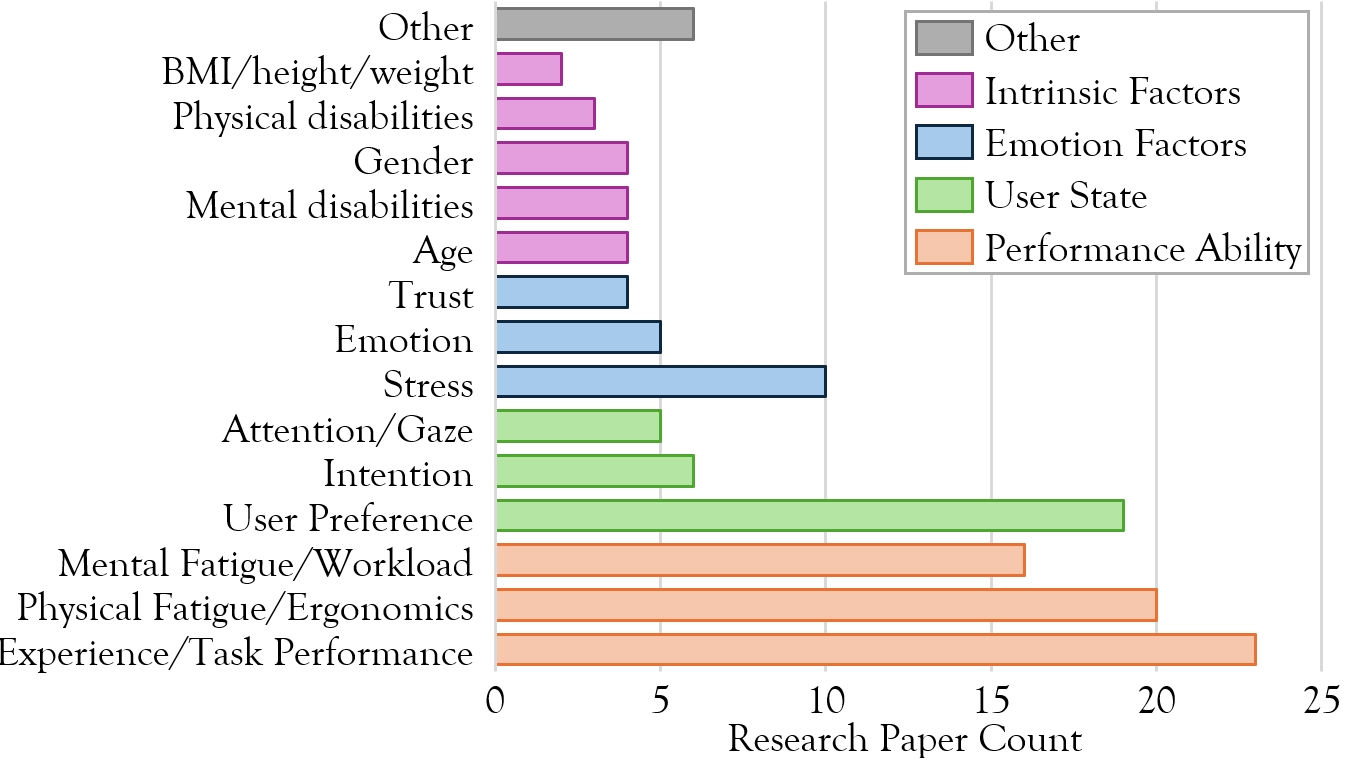}
    \vspace{-0.6cm}
    \caption{Analysis of key human factors HRC is adapted to.}
    \vspace{-0.4cm}
    \label{fig:personalisation}
\end{figure}

Literature analysis of the different personal factors used to customise HRC is shown in Fig.~\ref{fig:personalisation}. A large proportion of works focus on adaption based on a user's performance, such as their speed/experience, or their fatigue level. This coincides with the large proportion of works investigating task allocation and coordination, where the performance of a worker is used to inform how to optimise resources. User preference is a further key area of research, often aligned with task order preference, interface design or communication modality selection.

Human factor variation can occur between users (inter-user variation) and over time for a specific user (intra-user variation). Although the distinction between the two is not always clear, it presents a useful framework under which to discuss the various factors to which a robot may adapt. In the following subsections, key aspects of inter- and intra- user variation are highlighted, with examples of adaptive HRC.

\subsection{Inter-User Variation}
Inter-user variation encompasses differences between users that are predominantly static for a single user. Variation can take various forms and may be related to identity, physical or mental factors. Some key works researching HRC adaption based on inter-personal variation are discussed below.

\subsubsection{Identity Factors}
Identity factors include a user's cultural background, language, sex, age, etc. The increasing complexity of communication between robot and human in manufacturing, and the growing ability for speech/text recognition and comprehension with the use of Large Language Models (LLMs), is giving rise to language-based interfaces~\cite{dhanda_reviewing_2025, li_tod4ir_2022}. However, in a varied workforce consideration must be given to whether users can speak and interpret such commands sufficiently well, and whether other factors such as a user’s accent or dialect may prevent them from having their speech recognised. One of the few outputs acknowledging the different language needs of users is~\cite{freire_socially_2024}.

Non-verbal communication is an important method of communication but is not necessarily consistent across cultures. Facial expression recognition for use in HRC tasks has been investigated~\cite{chiurco_real-time_2022}, however there is a lack of acknowledgement for how results may change with cultural background which is known to have an impact~\cite{jack_facial_2012}. User preference for interaction modality, implicit or explicit communication and proximity to humans has been done with social robots, identifying key differences between cultures as well as significant variation within demographic groups~\cite{gasteiger_factors_2023}. There is therefore a need to investigate this further in industrial HRC settings.

Differences in perceptions of the benefits of HRC assistance by gender were investigated for a collaborative polishing task where high and low automation options were available. Males did not perceive high automation to provide a benefit while females did perceive a performance benefit~\cite{hopko_effect_2021}. The gender of workers has also been considered for assembly line balancing, considering an individual's comfortable working time~\cite{pabolu_development_2025} and ergonomics~\cite{mura_improving_2023}.

\subsubsection{Physical Factors}
Physical factors include a wide range of considerations. For example, height and weight, sensory dis/abilities such as hearing, sight, touch, and motor dis/abilities such as dexterity and mobility considerations. In I5.0 there is the potential to revolutionise the accessibility of jobs to those with physical impairment by integrating advanced interfaces such as generative AI chat tools and extended reality (XR) interfaces with robot manipulators. Supplementing a person’s physical limitations while making full use of their cognitive ability allows for a more inclusive workforce~\cite{rojas_lack_2024}. Some key works looking at HRC for physically disabled persons include mapping physical limitations to appropriate robot technology~\cite{lottermoser_method_2024}, and adapting interface modality for visually impaired users~\cite{stohr_adaptive_2018}. Key advantages of multimodal and adaptive communication methods are improving user experience through personalised choice, greater reliability through redundancy and increased accessibility. By developing multimodal communications as standard, persons with sensor impairments, e.g. deaf, blind, or those with communication limitations e.g. mute, limited dexterity, can still interact with a collaborative workcell. Adaption to physical factors is primarily through workcell and interface design, discussed further in Section~\ref{sec:workcell_interface}.

\subsubsection{Mental Factors}
Mental variation examples include learning speed, comprehension ability, concentration span or traits associated with neurodivergence e.g. dyslexia, ADHD and Autism Spectrum Disorder (ASD). A discussion on technology enabling the inclusion of disabled workers is presented in~\cite{rojas_lack_2024}, with some key highlights presented here.

Kildal et al. reduce the cognitive load of interpreting complex assembly instructions for workers unable to perform this step of the task. A robot gives specific instructions for where to connect wires to, allowing the user to perform dexterous tasks under the guidance of the robotic assistant~\cite{kildal_empowering_2019}. This is further developed with increased personalisation so workers with different abilities can choose the amount of assistance required from the robot counterpart on demand~\cite{kildal_collaborative_2021}. Adaption of collaborative assembly instructions for users with visual and/or cognitive impairment has been shown through adaptable displays, multimodal (touch, speech, audio) interfaces and the ability for users to switch mid-task promoting varied work tasks~\cite{stohr_adaptive_2018}. Analysis of differences in the behavioural patterns of neurotypical vs ASD operators during a HRC manufacturing task provided insights on factors such as task prioritisation, gaze, facial expressions, ability to adapt and performance, highlighting the need for adaptable interfaces for diverse users~\cite{mondellini_behavioral_2023}.

Mandischer et al. present a method for mapping various personal factors, such as ergonomics, motor function, cognition and body motion, along with how their abilities change over time (e.g. due to fatigue). The system can then allocate appropriate tasks to the user or robot for optimal completion. A key advantage is the applicability to people with disabilities, as well as able-bodied, elderly and injured people to adapt to the person's current working ability~\cite{mandischer_toward_2023}.

\subsection{Intra-User Variation}
Intra-user variation concerns the changing state of an individual user over time. These factors include the user's emotional state, stress, fatigue, experience level and task competence. Adaptive user-centred HRC requires real-time reaction to the user’s current physical and cognitive state.

\subsubsection{Stress}
For many users, working with cobots may be a new experience. This, combined with the dynamic nature of the technology, can lead to stress while working alongside cobots. Investigations have found users experience higher stress when approaching and programming a robot for the first time, although this reduces with time~\cite{lagomarsino_pick_2023, borghi_assessing_2025}.

Focussing on measuring stress objectively using electrocardiogram (ECG), the speed of a robot was adjusted to match that of a user in a collaborative assembly task. By correctly matching their speed, the method maintained high production levels while reducing the user's stress. A key point raised is that the user's experience of stress is subjective, therefore the objective measures used do not necessarily capture the full picture~\cite{ojstersek_personalizing_2024}. A similar work demonstrated the benefits of adjusting robot speed based on user stress and pace to maintain good productivity with reduced stress compared to no adjustment in robot speed or only adjusting based on the user's speed~\cite{messeri_human-robot_2021}.

\subsubsection{Trust}
Trust is key for effective HRC,
with a critical need for user trust in a system to be correctly calibrated to its ability. Otherwise, the system risks being used inappropriately due to over trust or unused due to lack of trust~\cite{gebru_review_2022}.

Investigating a user’s trust in a robot during tasks that were either high or low proximity, and high or low risk, showed the user had greater trust in low risk, low proximity situations~\cite{campagna_analysis_2023}. Further developments investigated adjusting the robot’s velocity, separation and proximity to the user’s head based on user trust. Users with HRC experience were happy to manually tune parameters while novice users preferred the system to automatically choose optimal parameters~\cite{campagna_promoting_2024}. This work highlights key points around the need to balance and optimise behaviour based on multiple, potentially conflicting parameters, and the varying level of user preference as to how much control they have over a robot’s behaviour.

\subsubsection{Fatigue}
Fatigue is a key consideration investigated in different works given its influence on productivity and user experience. Experiments with high and low levels of automation assistance in a collaborative polishing task found the high level of assistance helped enable fatigue recovery, however this reduces operator situational awareness~\cite{hopko_effect_2021}.
Future advanced HRC workcells may help increase situational awareness through sensing and intelligence~\cite{berx_harmonious_2025}.

\subsubsection{Emotion}
The recognition of a wide range of emotions for use in human state monitoring has been discussed in relation to smart manufacturing~\cite{wang_toward_2022} and the development of human digital twins~\cite{a_kiourtis_xr50_2024}. Emotion recognition methods include facial expression~\cite{chiurco_real-time_2022} and physiological data~\cite{eyam_emotion-driven_2021, gervasi_user_2022}.
Emotion recognition can be used to adapt a robot's behaviour, such as speed, location and delay, based on a user's current emotional state~\cite{eyam_emotion-driven_2021, gervasi_user_2022}.

A holistic review of the HRC experience highlights emotional intelligence as a key requirement, along with conversational skills, embodied intelligence, and the need to balance human autonomy with control~\cite{berx_harmonious_2025}. Limited discussion has been had on which emotions should be recognised and for what purpose in the context of HRC. Recognition of sensitive personal information by an employer has significant privacy, legal and ethical concerns, discussed further in Section~\ref{sec:ethics}.

\subsubsection{Cognitive Load Monitoring}
Various works have investigated monitoring cognitive load during HRC. The NASA-TLX questionnaire is a common method for assessing workload and has been used in HRC tasks to assess user workload under different robot conditions~\cite{hopko_effect_2021, javernik_nasa-tlx_2023}. For online workload monitoring, heart rate variability (HRV) has been used to increase robot assistance during high-stress conditions~\cite{landi_relieving_2018} and adjust the robot control parameters~\cite{lagomarsino_robot_2022}. A study looking at online and offline metrics found observability and transparency of robot actions impact user cognitive load, and validated that the proposed novel online metrics correlate well with standard offline metrics~\cite{lagomarsino_pick_2023}.

\subsection{Human Digital Twin}
Modelling a user’s various physical, mental and emotional traits, preferences and abilities leads to the concept of a Human Digital Twin (HDT)~\cite{wang_toward_2022, a_kiourtis_xr50_2024}. The use of the term HDT is not necessarily consistent across the literature, hence a cross-domain analysis was conducted and provides useful insights and definitions on what constitutes a HDT, human digital model and human digital shadow~\cite{gaffinet_human_2025}. In their definition, a HDT must provide a digital representation of a specific individual, with automatic data transfer from human to twin and twin to human.

The concept of digital twins for I5.0 in relation to extended reality (XR) technologies is discussed in detail by Kiourtis et al. The current status of “onesize-fits-all” XR solutions disregards the unique requirements and preferences of workers which should be considered for effective use in I5.0~\cite{a_kiourtis_xr50_2024}. The user model presented as part of the SHAREWORK project illustrates the use of a diverse model in a complete manufacturing task setup. A user’s physical, health, cognitive, expertise and performance states are used to inform HRC~\cite{umbrico_design_2022}. While some human factors can be monitored or modelled reliably with current technology, others present a significant challenge. Modelling cognitive state is an ambition presented in~\cite{gaffinet_human_2025}, though simpler HDTs can still provide benefits to HRC in many practical applications.

The overarching concept of HDT requires monitoring and storing of user characteristics over time, enabling predictions and insights to be gained. This raises significant ethical and regulatory questions that are frequently not considered in the research and will be discussed further in Section~\ref{sec:ethics}.

\section{HRC Workcell and Interaction Design}\label{sec:workcell_interface}
The new paradigm of humans and robots working together has necessitated developing spaces for effective HRC. Much of the research in this area takes a human-centric and task-efficiency approach. In this study however, we focus on works with a user-centric view, where aspects of the workcell and interaction are adjustable to individual users.



Fletcher et al. highlight the support of industry workers for ``the development of adaptive, flexible and self-optimising assembly systems of the future to meet both production and operator changes and differences, and the use of novel technologies". The work goes on to caveat this with the limited demographic of users surveyed, highlighting the need for broader studies investigating cultural differences across a range of demographics~\cite{fletcher_adaptive_2020}. Also highlighted is the desire for Augmented Reality (AR) and Virtual Reality (VR) as methods of delivering information, instructions and training. Various studies have developed systems with personalisable communication modalities and interface design. The multimodal, customisable nature of interfaces, including AR, a tablet and a smartwatch, were found to be key benefits in HRC manufacturing~\cite{andronas_multi-modal_2021}.

The level of instruction provided to a user should be adaptable based on their expertise and experience. This is demonstrated through the gradual removal of information from novices to experienced users in AR~\cite{umbrico_design_2022}. In a similar work, the rate of information decay during training using AR was adapted to the user~\cite{li_adaptive_2024}. This highlighted the importance of giving personalised information to the user for efficient learning and the complexity of doing this in practice, with a need for further investigations on broader populations.


Personalised gesture meaning for robot communication has been shown to reduce physical and mental workload~\cite{rautiainen_multimodal_2022}. The work also highlights the need for workers to move between workstations, hence having adaptable interfaces reduces the need for users to learn different systems. The growing prevalence of generative AI language tools presents a key shift in possibility for human-robot communication~\cite{dhanda_reviewing_2025}. The GPT based dialogue tool ToD4IR is significant given its ability for practical task-oriented discussion and natural, small talk style language~\cite{li_tod4ir_2022}. Being able to converse naturally with a robot may help unskilled users interact confidently, a task which may otherwise be stressful~\cite{borghi_assessing_2025}.


\section{Collaborative Task Completion}\label{sec:task_completion}
A further area for personal differences in HRC manufacturing is how a task progresses to be completed. For a given task with multiple subtasks, there is significant scope for variation in which agent performs which actions, in what order the actions are performed, and how long a specific agent may take to do a task.

\subsection{Task Allocation}
Various works have investigated the assignment of actions to a robot or human during the planning of a task based on their respective capabilities. These methods do not necessarily account for the individual characteristics and abilities of unique workers, hence the subsequent examples focus on investigations that consider solutions tailored to individuals.

The need for task allocation to take into account multiple factors including ``productivity, flexibility, and human factors" is investigated in~\cite{calzavara_multi-objective_2023}, where a multiobjective action allocation optimisation method considers makespan, human energy expenditure and mental workload. Subsequent developments include a real-time dynamic system~\cite{calzavara_achieving_2024} and individualisation to workers, with personalised rest allocation based on age, bodyweight and gender used to help balance the assembly line~\cite{keshvarparast_integrating_2024}.

Task scheduling optimisation in a multi-user, HRC task using personalised fatigue models was investigated to jointly optimise for cycle time while minimising team fatigue or an individual’s fatigue~\cite{chand_dual_2023}. A motivating factor for the work is promoting fairness of job allocation between workers, however this raises the need for consideration as to what constitutes fair job allocation. 
A genetic algorithm method balanced worker energy expenditure and noise exposure for a task rotation plan, with individual characteristics (age, weight, height, gender, skill level and cost) considered in the plan, creating a more ergonomic work environment~\cite{mura_improving_2023}.

A key work using dynamic fatigue assessment and task allocation is~\cite{you_human_2025}, however the work requires further development to optimise for multiple factors. The benefits of adaptive task allocation in HRC to maintain well-being and motivation with reduced monotony are discussed in detail~\cite{berx_harmonious_2025}, though few works have investigated adaptive planning with this as the focus.

\subsection{Task Timing}
Users perform actions at different speeds, and that speed will vary over time based on factors such as experience and fatigue. For optimal synchronisation with a user, it is beneficial to model prior expectations of how long they will take on a task (e.g. based on previous experience), and also adapt based on real-time monitoring of task progression.

Adaptive action timing prediction has been investigated based on current action recognition and adjustment based on previous experience from the user~\cite{male_deep_2023}, enabling synchronisation of collaborative actions. A further example of time planning uses Bayesian Reinforcement Learning (RL) to adjust the speed of a robot feeding an assembly line. The method adjusts the robot speed to match the speed of different workers on different tasks, balancing the assembly line and helping prevent workers from feeling overloaded~\cite{lee_bayesian_2024}. Action timeline planning while accounting for duration uncertainty is investigated with the ROXANNE planner, where integrating a personalised duration expectation and expected variance in duration based on worker experience level helps create an adaptive task planner for collaborative tasks~\cite{umbrico_design_2022}.

\subsection{User Preference}
Users may exhibit preference as to how a task should progress towards completion. For example, what order actions occur in~\cite{obidat_development_2023}, or on which tasks they would like assistance. Investigations found workers have a preference for having input on task allocation decisions, rather than relying on a foreman or automated system to decide for them~\cite{tausch_human-robot_2022}. Similar findings were reported in~\cite{c_schmidbauer_empirical_2023}, which also found workers tend to prefer allocating physical tasks to robots while keeping cognitive tasks for themselves.

The real-time adaption of robot behaviour using a partially observable Markov model allows for dynamic replanning, error recovery and adaption to user preference. The method allows a robot to decide when to retrieve parts/tools, clear the workspace or assist the user by holding a part, while implicitly learning the user’s preference on-the-fly~\cite{mangin_how_2022}.


\section{Ethical and Safety Considerations}\label{sec:ethics}
The benefits of personalised HRC are numerous and include improved well-being, efficiency, safety and inclusivity of a varied workforce. However, personalised HRC also raises various ethical and safety considerations which are not discussed substantially in the research~\cite{sumak_sensors_2022}. While a review of safety in collaborative intelligence is available~\cite{ramos_collaborative_2024}, this section highlights and builds on particularly salient points for personalised HRC in manufacturing.

Control strategies that adapt to different users present the opportunity for improved user perception and greater safety through real-time collision avoidance. However, adaptive and unpredictable movements may lead to reduced trust and decreased safety if errors occur. The variability of workers, uncertainty in human behaviour and probabilistic AI methods present challenges when ensuring safe and consistent performance~\cite{ramos_collaborative_2024}. Novel risk assessment methods are required in light of adaptable and dynamic robotic systems~\cite{umbrico_design_2022}.

Discussing the data privacy of workers considering highly personalised systems is crucial. Insights from surveys of manufacturing workers highlight the desire for personnel to have access to their personal data and the need for strict data security and protection~\cite{fletcher_adaptive_2020}.
In this emerging field, various regulations need to be considered and reviewed as technologies develop. In Europe, relevant regulations include the General Data Protection Regulation (GDPR)~\cite{EU_GDPR_reg_2016}, the EU Machinery Regulation~\cite{EU_machinery_reg_2023} , and the EU AI Act~\cite{EU_AI_reg_2024}.

The Machinery Regulation (EU) 2023/1230 now includes reference to machines with self-evolving behaviour in relation to control and ergonomics. Guidance on reducing operator ``discomfort, fatigue and physical and psychological stress" highlights the use of machines which allow variable user pose, avoid imposed rates of work and use self-evolving behaviour for different communication modalities, including transparency of actions~\cite{EU_machinery_reg_2023}.
The EU AI Act presents guidance on a range of relevant points. Given the unreliability and intrusiveness of recognising emotion from biometric data, restriction is given to prohibit the recognition of emotion in workplace settings. This includes ``happiness, sadness, anger, [etc.]" but does not include physical states such as pain and fatigue. Included under the high-risk category is the use of AI ``to allocate tasks based on individual behaviour or personal traits or characteristics or to monitor and evaluate the performance and behaviour of persons in such relationships"~\cite{EU_AI_reg_2024}. Various challenges related to AI implementation in manufacturing still exist, such as security, feedback integration, explainability, accountability, and uncertainty awareness~\cite{li_trustworthy_2025}.

GDPR provides various regulations on the use of personal data in the workplace, particularly around profiling employees with respect to sensitive information~\cite{EU_GDPR_reg_2016}. While the storing and use of personal data are protected, an employee may feel pressured by an employer to consent to the use of their data in a personalised HRC setting where greater efficiencies can be achieved. The significant ethical and security considerations for storing personal data in the context of HDTs is an area needing further research~\cite{gaffinet_human_2025}.
Particularly for task allocation and scheduling, various research outputs are increasingly using highly personal characteristics. For example, using age, gender and BMI to design work schedules based on predicted comfortable working duration, with explicit prioritisation of older workers~\cite{pabolu_development_2025}, or rest allocation based on age, weight and gender~\cite{keshvarparast_integrating_2024}. Although well intentioned, these methods present significant opportunities for profiling workers based on protected characteristics which should be carefully considered in future developments.

The development over recent years of regulations that consider the use of AI, personal data and intelligent machines has helped provide guidance on how such technologies may be used, particularly relevant in the case of personalised HRC in industry. As technologies are developed and brought to market, further discussion and updates to regulation will be required. Particularly in the EU, the desire for a user-centric I5.0~\cite{directorate-general_EC_industry50_2021} may align in vision with the protections and guidelines of other regulations~\cite{EU_GDPR_reg_2016, EU_machinery_reg_2023, EU_AI_reg_2024}, but in practice there may be tensions in deploying new technologies.

\section{Discussion on Future Research Directions}\label{sec:future_work}

One of the clearest needs for future developments is improved experimentation through a broad range of representative participants with rigorous methods and analysis. The validity of increasingly personalised interactions can only be shown through testing on a range of participants, an area in which many research outputs lack rigour. This requires greater cross-disciplinary integration, where experiments are conducted in collaboration with experts in biomechanics, psychology, human factors, etc. This is linked to the lack of research focussing on consistency previously identified~\cite{faccio_human_2023}.

Research on increased accessibility of HRC for users with disabilities has started, though there is significant room for further development~\cite{rojas_lack_2024, sumak_sensors_2022}. Instead of specialising HRC platforms for specific disabilities, a broader focus on highly adaptive, multimodal interfaces allowing accessibility to a broad range of users may be more impactful. This promotes the inclusion of a wide spectrum of users, including those with and without disabilities, and encourages the expectation of user-adaptable systems, as highlighted in~\cite{mandischer_toward_2023}.

Various works have been shown to help reduce the negative physical and mental impacts of collaborative manufacturing tasks through adaptive behaviour. Further developments may take a proactive role in promoting user well-being, for example through prompting regular (active) breaks and engaging a user’s mental capacities by providing levels of autonomy~\cite{berx_harmonious_2025} and stimulation throughout a task.

Monitoring of various physical and mental parameters, such as ergonomics, emotion, working speed, and preferences, have been discussed in the literature for adaptable robot control. However, there lacks a rigorous and consistent methodology for modelling a HDT~\cite{gaffinet_human_2025}. Developing a unified approach would enable greater collaborative research and present a grounded framework from which ethical and regulatory questions can be discussed. The parameters modelled should have clear reasons for being monitored, and such a system must ensure high levels of inbuilt security to protect workers’ privacy.

Surprisingly, it is foreseeable that some of the technologies discussed could lead to a reduced user-centric focus. While the technologies are developed to increase well-being and promote customised interactions based on individual needs, the opposite effect could be realised. From a business perspective, by transferring the burden of considering individual needs from managers to technology, employees may lose their individuality. This is seen in the development of methods to adapt task scheduling according to individual ability and personal characteristics~\cite{pabolu_development_2025, mura_improving_2023, keshvarparast_integrating_2024}. In such scenarios, employees have limited opportunity to work according to their subjective experience as the system has dictated their ability, and it is assumed this gives a fair assessment of their current capacity. Such arguments may be speculative, but consideration should be given to ensure human needs beyond the scope of technological developments are accounted for.

Unification of metrics for assessing the suitability of personalised HRC technologies are required for reliable and comparative results. A range of methods are currently used in research, primarily falling into the categories of business (e.g. cycle rates, completion times, cost effectiveness, etc.), human factors (e.g. user acceptance, stress, fatigue, ergonomics, etc.) and technological (e.g. accuracy, error rate, latency, etc.). Evaluating relevant factors across the three categories should be normalised, with an emphasis on increasing the analysis of user perspectives towards new technological developments. The need for increased analysis of individual experiences in HRC is also highlighted in~\cite{vermeulen_safety_2024}, where a conceptual framework for examining human factors is presented.

Future developments will lead to an increased variety of robot behaviours which can be comprehensively adapted. Personalisation for user experience, proactivity, multi-user engagement and generalisation across diverse and new tasks may be included. These developments will coincide with a greater focus on  diverse robot platforms, such as mobile bases, humanoids and quadrupeds, giving added personalisation opportunities not possible with table-mounted cobot arms used in the majority of current research.


\section{Conclusion}\label{sec:conclusion}
There has been a growing trend of personalised HRC in manufacturing over recent years, in part due to the ambitions of Industry 5.0. This review has shed light on key aspects of personalised HRC, including which personal factors are used, how robot interfaces and experiences are customised and how collaborative task completion can be adapted to individuals. There has been a further discussion on ethical and regulatory considerations, which are often not fully discussed in the literature. Further more, key avenues of future research and associated guidance are presented in what promises to be a growing field of research.

\section*{Acknowledgements}
This project has received funding from the European Union's Horizon Europe research and innovation programme under Grant Agreement no. 101059903 and 101135708. 

\bibliographystyle{IEEEtran}
\bibliography{IEEEabrv,reference.bib}

\end{document}